\documentclass[sigconf]{acmart}

\usepackage{booktabs} 
\usepackage{color}

\usepackage[utf8]{inputenc} 
\usepackage[T1]{fontenc}    
\usepackage{hyperref}       
\usepackage{url}            
\usepackage{booktabs}       
\usepackage{amsfonts}       
\usepackage{nicefrac}       
\usepackage{microtype}      
\usepackage{algorithm,algorithmicx,algpseudocode}
\usepackage{amsmath}
\usepackage{graphicx}
\usepackage{subfig}
\usepackage{bm}
\usepackage{flushend}
\usepackage{wrapfig}
\usepackage{wrapfig,lipsum,booktabs} 
\usepackage{multirow}
\usepackage[normalem]{ulem}
\usepackage{soul}
\usepackage{enumitem}
\DeclareMathOperator*{\argmin}{arg\,min} 
\DeclareMathOperator*{\relu}{ReLU}
\newcommand{\IR}{\mathbb R}
\newcommand{\II}{\mathbb I}

\usepackage{breqn}

\acmDOI{10.475/123_4}

\acmISBN{123-4567-24-567/08/06}

\begin{document}
\title{Subspace Network: Deep Multi-Task Censored Regression for Modeling Neurodegenerative Diseases}

\author{Mengying Sun$^1$, Inci M. Baytas$^1$, Liang Zhan$^2$, Zhangyang Wang$^3$, Jiayu Zhou$^1$}
\affiliation{%
  \institution{$^1$Computer Science and Engineering, Michigan State University\\
               $^2$Computer Engineering Program, University of Wisconsin-Stout\\
               $^3$Computer Science and Engineering, Texas A\&M University
               }
}
\email{{sunmeng2, baytasin}@msu.edu, zhanl@uwstout.edu, atlaswang@tamu.edu, jiayuz@msu.edu}



\renewcommand{\shortauthors}{M. Sun et al.}
\renewcommand{\shorttitle}{Subspace Network: Deep Multi-Task Censored Regression}

\begin{abstract}
Over the past decade a wide spectrum of machine learning models have been 
developed to model the neurodegenerative diseases, associating biomarkers, especially
non-intrusive neuroimaging markers, with key clinical scores measuring
the cognitive status of patients. Multi-task learning (MTL) has been commonly utilized by these studies to address high dimensionality and small cohort size challenges. However, most existing MTL 
approaches are based on linear models and suffer from two major limitations: 1) they cannot
explicitly consider upper/lower bounds in these clinical scores; 2) they lack the capability to capture complicated non-linear interactions among the variables. In this paper, we propose {\it Subspace Network}, an efficient
deep modeling approach for non-linear multi-task censored regression. Each
layer of the subspace network performs a multi-task censored regression to
improve upon the predictions from the last layer via sketching a 
low-dimensional subspace to perform knowledge transfer among learning tasks. Under mild assumptions, for each layer the parametric
subspace can be recovered using only one pass of training data. Empirical results demonstrate that the proposed subspace network quickly picks up 
the correct parameter subspaces, and outperforms state-of-the-arts in predicting 
neurodegenerative clinical scores using information in brain imaging. 
\end{abstract}

%


\maketitle

\section{Introduction}\label{sec:intro}


Recent years have witnessed increasing interests on applying machine learning 
(ML)
techniques to analyze biomedical data. Such data-driven approaches deliver promising performance improvements in many challenging
predictive problems. For example, in the field of neurodegenerative diseases
such as Alzheimer's disease and Parkinson's disease, researchers have
exploited ML algorithms to predict the cognitive functionality
of the patients from the brain imaging scans, e.g., using the magnetic resonance
imaging (MRI) as in \cite{adeli2015robust,zhang2012multi,zhou2011multi}. A key finding points out that there are typically various types of prediction 
targets 
(e.g., cognitive scores), and they can be jointly learned 
using multi-task 
learning (MTL), e.g., ~\cite{caruana1998multitask,evgeniou2004regularized,zhang2012multi}, 
where the predictive information is shared and transferred among related models 
to reinforce their generalization performance. 

\begin{figure*}[ht]
	\vspace{-1em}
	\centering
	\includegraphics[width=0.9\textwidth]{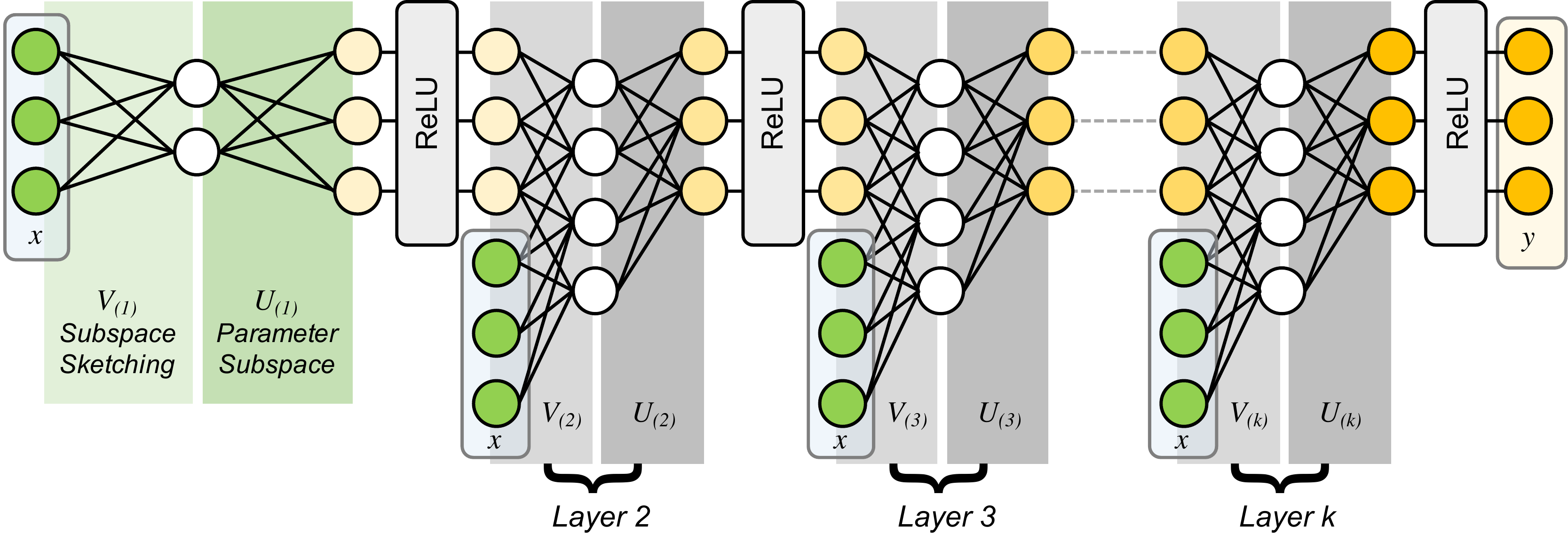}
	\vspace{-0.2in} 
	\caption{The proposed subspace network via hierarchical 
		subspace sketching and refinement. }
	\label{fig:network_structure}
	\vspace{-1.5em}
\end{figure*}

Two challenges persist despite the progress of applying MTL in disease
modeling problems. First, it is important to notice that
 clinical targets, different from
typical regression targets, are often naturally bounded. 
For example, in the output of Mini-Mental State Examination (MMSE) test, a key reference for deciding cognitive impairments, ranges from 0 to
30 (a healthy subject): a
smaller score indicates a higher level of cognitive dysfunction (please refer to~\cite{tombaugh1992mini}). Other
cognitive scores, such as Clinical Dementia Rating Scale
(CDR)~\cite{hughes1982new} and Alzheimer's Disease Assessment Scale-Cog (ADAS-
Cog)~\cite{rosen1984new}, also have specific upper and lower bounds. Most existing approaches, e.g.,~\cite{zhang2012multi,zhou2011multi,poulin2011amygdala}, relied on linear regression without
considering the range constraint, partially due to the fact that mainstream MTL models for regression, e.g., \cite{jalali2010dirty,argyriou2007multi,zhang2012multi,zhou2011malsar}, are developed using the least squares 
loss and cannot be directly extended to censored regressions. As the second challenge, a majority of MTL research focused on linear models because of computational efficiency and theoretical guarantees. 
However, linear models cannot capture the complicated non-linear relationship between features and clinical targets. 
For example, \cite{alzheimer20132013} showed the early onset of Alzheimer's disease to be related to  single-gene mutations on chromosomes \textit{21}, \textit{14}, and  \textit{1}, and the effects of such mutations on the cognitive impairment are hardly linear (please refer to \cite{martins2005apoe,sweet2012effect}). Recent advances in multi-task deep neural networks~\cite{seltzer2013multi,zhang2014facial,wu2015deep} provide a promising direction, but their model complexity and demands of huge number of training samples prohibit their broader usages in clinical cohort studies.

To address the aforementioned challenges, we propose a \textit{novel and efficient} deep
modeling approach for non-linear multi-task censored regression, called
\textit{Subspace Network} (SN), highlighting the following multi-fold technical innovations:
\vspace{-0.5em}
\begin{itemize}[leftmargin=0.1in]
\setlength\itemsep{0.1em}
\item It efficiently builds up a deep network in a  \textbf{layer-by-layer feedforward} fashion, and in each layer considers a \textbf{censored} regression problem. The layer-wise training allows us to grow a deep model efficiently.
\item It explores a \textbf{low-rank} subspace structure that captures task relatedness for better predictions. A critical difference on subspace decoupling between previous studies such as \cite{mardani2015subspace}  \cite{shen2016online} and our method lies on our assumption of a low-rank structure in the parameter space among tasks rather than the original feature space. 
\item By leveraging the recent advances in online subspace sensing~\cite{mardani2015subspace,shen2016online}, we show that the parametric 
subspace can be recovered for each layer with feeding \textbf{only one pass} of
the training data, which allows more efficient layer-wise training. 
\end{itemize}
\vspace{-0.5em}
Synthetic experiments verify the technical claims of the proposed SN, and it outperforms various state-of-the-arts methods in modeling neurodegenerative diseases on real datasets.


\section{Multi-task censored regression via parameter subspace sketching and refinement}
\label{sec:method:onelayer}


In censored regression, we are given a set of $N$ observations $\mathcal D=
\{(x_i, y_i)\}_{i=1}^N$ of $D$ dimensional feature vectors $\{x_i \in
\IR^D\}$ and $T$ corresponding outcomes $\{y_i\in \IR^{T}_+\}$, where each
outcome $y_{i,t}\in \IR_+$, $t \in \{1,\cdots,T \}$, can be cognitive scores 
(e.g., MMSE and ADAS-Cog)
or other biomarkers of interest such as proteomics\footnote{Without loss of
generality, in this paper we assume that outcomes are lower censored at 0. By
using variants of Tobit models, e.g., as in~\cite{shen2016online}, the proposed algorithms
and analysis can be extended to other censored models with minor changes in
the loss function.}. For each outcome, the censored regression assumes a nonlinear
relationship between the features and the outcome through a rectified linear unit 
(ReLU) transformation, i.e., $y_{i, t} = \relu\left(W_t^\top x_i +
 \epsilon\right)$ where $W_t\in \IR^D$ is the coefficient for input features, $\epsilon$ is 
{\it
i.i.d.} noise, and ReLU is defined by $\relu (z) = \max (z, 0)$.
We can thus collectively represent the censored regression for multiple tasks
by:
\begin{align}
y_i = \relu \left(W x_i + \epsilon\right), \label{eq:onelayer_model}
\end{align}
where $W = [W_1, \dots, W_T]^\top \in \IR^{T \times D}$ is the coefficient
matrix. We consider the regression problem
for each outcome as a learning task. One commonly used task relationship
assumption is that the transformation matrix $W \in \IR^{T \times D}$ belongs
to a linear low-rank subspace $\mathcal U$. The subspace allows us to
represent $W$ as product of two matrices, $W = UV$, where columns of $U\in 
\IR^{T \times R}= [U_1, \dots,
U_T]^\top$ span the linear subspace $\mathcal U$, and $V\in \IR^{R
\times D}$ is the embedding coefficient. We note that the output $y$ can be
entry-wise decoupled, such that for each component $y_{i,t} = \relu(U_t^{\top} V 
x_i +
\epsilon)$. By assuming Gaussian noise $\epsilon \sim \mathcal N (0,
\sigma^2)$, we derive the following likelihood function:
\begin{align*}
\Pr(y_{i, t}, x_i | U_t, V) = 
   \phi \left( \frac{y_{i, t} - U_t^\top V x_i}{\sigma}\right) \II(y_{i, t} \in 
(0, \infty)) \\
	+ \left[1- Q \left( \frac{0 - U_t^\top V x_i}{ \sigma} \right)\right] 
\II(y_{i, t} = 0),
\end{align*}
where $\phi$ is the probabilistic density function of the standardized
Gaussian $N(0,1)$ and $Q$ is the standard Gaussian tail. $\sigma$ controls how accurately the low-rank subspace assumption can fit the data. Note that other noise models can be assumed here as well. The likelihood of
$(x_i, y_i)$ pair is thus given by:
\begin{align*}
\Pr(y_i, x_i | U, V) = 
   \prod_{t=1}^T \bigg \{ \phi \left( \frac{y_{i, t} - U_t^\top V 
x_i}{\sigma}\right) \II(y_{i,t} \in (0, \infty)) \\
	+ \left[1- Q \left( -\frac{U_t^\top V x_i}{ \sigma} \right)\right] \II(y_{i, t} 
= 0) \bigg \}.
\end{align*}
The likelihood function allows us to estimate subspace $U$ and coefficient $V$
from data $\mathcal D$. To enforce a low-rank subspace, one common approach is
to impose a trace norm on $UV$, where trace norm of a matrix $A$ is defined by $\|A\|_* = \sum_j s_j (A)$ 
and $s_j(A)$
is the $j$th singular value of $A$. Since $\|UV\|_* =
\min_{U,V}\frac{1}{2}(\|U\|_F^2 + 
\|V\|_F^2)$, e.g., see~\cite{srebro2005maximum,mardani2015subspace}, the objective function 
of
multi-task censored regression problem is given by: 
\begin{align}  
 \min\nolimits_{U, V} -\sum\nolimits_{i=1}^N\log\Pr(y_i, x_i | U, V) 
   + \tfrac{\lambda}{2} (\|U\|_F^2 + \|V\|_F^2). 
 \label{object}
\end{align}

\vspace{+0.2em}
\subsection{An online algorithm}

We propose to solve the objective in \eqref{object} via the block
coordinate descent approach which is reduced to iteratively updating the following two subproblems:
\begin{align}
V^+ &= \argmin\nolimits_{V} -\sum\nolimits_{i=1}^N\log\Pr(y_i, x_i | U^-, V)
   + \tfrac{\lambda}{2} \|V\|_F^2, \tag{P:V}\label{eq:solveV}\\
U^+ &= \argmin\nolimits_{U} -\sum\nolimits_{i=1}^N\log\Pr(y_i, x_i | U, V^+) 
   + \tfrac{\lambda}{2} \|U\|_F^2 \tag{P:U}.\label{eq:solveU}
\end{align}
Define the instantaneous cost of the $i$-th datum: 
\begin{align*}
g(x_i, y_i, U, V)  =  - \log\Pr(x_i, y_i | U, V) +  \tfrac{\lambda}{2}\|U\|_F^2 + 
\tfrac{\lambda}{2}\|V\|_F^2,
\end{align*} 
and the online optimization form of (\ref{object}) can be recast as an empirical cost 
minimization given below:
\begin{align*}
\min\nolimits_{U, V} \tfrac{1}{N}\sum\nolimits_{i=1}^N g(x_i, y_i, U, V).
\end{align*}
According to the analysis in Section~\ref{sec:method:onelayer:analysis},
one pass of the training data can warrant the subspace learning problem.
We outline the solver for each subproblem as follows:

\begin{algorithm}[!t]
\caption{Single-layer parameter subspace sketching and refinement.}
\begin{algorithmic}
\Require Training data $\mathcal D = \{(x_i, y_i)\}_{i=1}^N$, rank parameters $\lambda$ and $R$,  
\Ensure parameter subspace $U$, parameter sketch $V$
\State Initialize $U^-$ at random
\For {$i = 1, \dots, N$}
   \State \textit{// 1. Sketching parameters in the current subspace}
	   \begin{align*}
	   V^+ = \argmin\nolimits_{V} -\log\Pr(y_i, x_i | U^-, V)
   + \tfrac{\lambda}{2} \|V\|_F^2 
	   \end{align*}
   \State \textit{// 2. Parallel subspace refinement $\{U_t\}_{t=1}^T$}
   \For {$t = 1, \dots, T$}
   	\State $U_t^+ = \argmin\nolimits_{U_t} -\log\Pr(y_{i, t}, x_i | U_t, V^+)
   + \frac{\lambda}{2} \|U_t\|_2^2$
   \EndFor
   \State Set $U^- = U^+, V^- = V^+$
\EndFor
\end{algorithmic}
\label{alg:onelayer}
\end{algorithm}

\noindent{\bf Problem \eqref{eq:solveV} sketches parameters in the current space.}
We solve \eqref{eq:solveV} using gradient descent. The parameter sketching 
couples all the subspace dimensions in $V$ (not decoupled as
in~\cite{shen2016online}), and thus we need to solve this collectively. The 
update of $V$ ($V^+$) can be obtained by solving the online problem given below:
\begin{align*}
&\min_{V} g(x_i, y_i; U^{-}, V)
   \equiv  -\sum\nolimits_{t=1}^{T} \log \Pr(y_{i,t}, x | U^{-}_{t}, V) + 
\frac{\lambda}{2} \|V\|_F^2\\
&   = -\sum_{t=1}^{T} \log\bigg[\phi \left( \frac{y_{i,t} - 
\left(U_t^-\right)^\top V x}{\sigma}\right) \II(y_{i,t} \in (0, \infty))\\
& + \left[1- Q \left( \frac{-\left(U_t^-\right)^\top V x}{ \sigma} 
\right)\right] \II(y_{i,t} = 0)\bigg] + \frac{\lambda}{2} \|V\|_F^2.
\end{align*}
$V^+$ can be computed by the following gradient update:
$
V^+ = V^- - \eta \nabla_{V} g(x_i, y_i; U^-, V^+),
$
where the gradient is given by:
\begin{align*}
\nabla_{V} & g(x_i, y_i; U^-, V^+) =  \lambda V+ \sum_{t=1}^{T}\begin{cases}
- \frac{y_{i,t} - \left(U^{-}_t\right)^\top V x_i}{\sigma^2} U^{-}_t x_i^\top & 
y_{i,t} \in (0, \infty)\\
\frac{\phi(z_t)}{\sigma \left[1 - Q(z_{i,t})\right]} U^{-}_t x_i^T  & y_{i,t} = 
0
\end{cases}
\end{align*} 
where $z_{i, t} = \sigma^{-1} (- \left(U_{t}^-\right)^\top V x)$. 
The algorithm for solving \eqref{eq:solveV} is summarized in Alg.~\ref{alg:PV}.

\begin{algorithm}[!h]
	\caption{Gradient descent algorithm for problem~\ref{eq:solveV}.}
	\begin{algorithmic}
		\Require Training data $(x_i, y_i)$, $U^-$, step size $\eta$,  
		\Ensure sketch $V$
		\State Initialize $V^-$ at random.
		\State \textit{// 1. Perform gradient step and update the current solution of V}.
		\For {$t = 1, \dots, T$}
		\State Compute $z_{i,t} = \sigma^{-1} (- \left(U_{t}^-\right)^\top V x_i)$.
		\State Compute the gradient for $y_{t}$:
		\begin{align*}
		\nabla{g}_{t}&(x_i, y_{i,t}; U^-, V^+) = 
		\begin{cases}
		- \frac{y_{i,t} - \left(U^{-}_t\right)^\top V x_i}
		{\sigma^2} U^{-}_t x_i^\top & y_{i,t} \in (0, \infty)\\
		\frac{\phi(z_{i,t})}{\sigma \left[1 - Q(z_{i,t})\right]}
		U^{-}_{t} x_i^\top  & y_{i,t} = 0
		\end{cases}
		\end{align*}
		\EndFor
		\State \textit{// 2. Update the current sketch $V^-$}
		\begin{align*}
		V^+ = V^- - \eta \left[\sum\nolimits_{t=1}^{T} \nabla{g}_{t}(x, y_t; U^-, V^+)+ \lambda V\right]
		\end{align*}
		\State Set $V^- = V^+$
	\end{algorithmic}
	\label{alg:PV}
\end{algorithm}  


\noindent{\bf Problem \eqref{eq:solveU} refines the subspace $U^+$ based on sketching.} 
We solve \eqref{eq:solveU} using stochastic gradient descent (SGD). We note that 
the problem is decoupled for different subspace dimensions $t = 1, \dots, T$ 
(i.e., rows of $U$). With careful parallel design, this procedure can be done 
very efficiently. Given a training data point $(x_i, y_i)$, the problem related 
to the $t$-th subspace basis is: 
\begin{align*}
& \min_{U_t}  g_t(x_i, y_{i,t}; U_t, V^+) 
    \equiv -\log \Pr(y_{i,t}, x_i | U_t, V^+) + \frac{\lambda}{2} \|U_t\|_2^2\\
&  = -\log\bigg[\phi \left( \frac{y_{i,t} - U_t^\top V^+ x_i}{\sigma}\right) 
\II(y_{i,t} \in (0, \infty))\\
&  + \left[1- Q \left( \frac{ - U_t^\top V^+ x_i}{ \sigma} \right)\right] 
\II(y_{i,t} = 0)\bigg] + \frac{\lambda}{2} \|U_t\|_2^2.
\end{align*}
We can revise subspace by the following gradient update:
$
U_t^+ = U_t^- - \mu_t \nabla_{U_t} g_t(x_i, y_{i,t}; U_t, V^+),
$
where the gradient is given by:
\begin{align*}
\nabla_{U_t} &g_t(x_i, y_{i,t}; U_{i,t}, V^+)
= \lambda U_t + \begin{cases}
- \frac{y_{i,t} - U_t^\top V^+ x}{\sigma^2} V^+ x_i  & y_{i,t} \in (0, \infty)\\
\frac{\phi(z_{i,t})}{\sigma \left[1 - Q(z_{i,t})\right]} V^+ x_i & y_{i,t} = 0
\end{cases}
\end{align*}
where $z_{i,t} = \sigma^{-1} (- U_t^\top V^+ x_i)$. We summarize the procedure
in Algorithm~\ref{alg:onelayer} and show in
Section~\ref{sec:method:onelayer:analysis} that under mild assumptions this
procedure will be able to capture the underlying subspace structure in the
parameter space with just one pass of the data.

\vspace{+0.2em}
\subsection{Theoretical results}\label{sec:method:onelayer:analysis}


We establish both asymptotic and non-asymptotic convergence properties for 
Algorithm \ref{alg:onelayer}. The proof scheme is inspired by a series of 
previous works:~\cite{mairal2010online, kasiviswanathan2012online, 
shalev2012online, mardani2013dynamic, mardani2015subspace, shen2016online}. We 
briefly present the proof sketch, and more proof details can be found in Appendix. At each iteration $i = 1, 2,..., N$, 
we sample $(x_i, y_i)$, and let $U^i, V^i$ denote the
intermediate $U$ and $V$, to be differentiated from $U_t,
V_t$ which are the $t$-th columns of $U, V$. For the proof feasibility, we
assume that $\{(x_i, y_i)\}_{i=1}^N$ are sampled i.i.d., and the subspace
sequence $\{U^i\}_{i=1}^N$ lies in a compact set.

{\bf Asymptotic Case:}
To estimate $U$, the Stochastic Gradient Descent (SGD) iterations can be seen as 
minimizing the approximate cost $\frac{1}{N}\sum_{i=1}^N g'(x_i, y_i, U, V)$, 
where $g'$ is a tight quadratic surrogate for $g$ based on the second-order 
Taylor approximation around $U^{N-1}$. Furthermore, $g$ can be shown to be 
smooth, by bounding its first-order and second-order gradients w.r.t. each $U_t$ 
(similar to Appendix 1 of~\cite{shen2016online}). 

Following \cite{mairal2010online, mardani2015subspace}, it can then be established that, as $N \rightarrow \infty$, \textit{the subspace 
sequence $\{U^i\}_{i=1}^N$ asymptotically converges to a stationary-point of the 
batch estimator, under a few mild 
conditions}. We can sequentially show: 1) $\sum_{i=1}^N g'(x_i, y_i, U^i, V^i)$ asymptotically 
converges to $\sum_{i=1}^N g(x_i, y_i, U^i, V^i)$, according to the 
quasi-martingale property in the almost sure sense, owing to the tightness of 
$g'$; 2) the first point implies convergence of the associated gradient 
sequence, due to the regularity of $g$; 3) $g_t(x_i, y_i, U, V)$ is bi-convex 
for block variables $U_t$  and $V$.

{\bf Non-Asymptotic Case:}
When $N$ is finite,~\cite{mardani2013dynamic} asserts that \textit{the distance 
between successive subspace estimates will vanish as fast as $O(1/i)$: $\|U^i - 
U^{i-1}\|_F \le \frac{B}{i}$, for some constant $B$ that is independent of $i$ 
and $N$}. Following~\cite{shen2016online} to leverage the unsupervised 
formulation of regret analysis as in~\cite{kasiviswanathan2012online, 
shalev2012online}, we can similarly obtain a tight regret bound that will again vanish if 
$N \rightarrow \infty$.


\vspace{+0.2em}
\section{Subspace network via hierarchical sketching and refinement}
\label{sec:method:hierarchical}

\begin{algorithm}[!t]
	\caption{Network expansion via hierarchical parameter subspace sketching and refinement.}
	\begin{algorithmic}
		\Require Training data $\mathcal D = \{(x_i, y_i)\}$, target network depth $K$. 
		\Ensure The deep subspace network $f$
		\State Set $f_{[0]}(x) = y$ and solve $f_{[0]}$ using Algorithm \ref{alg:onelayer}.
		\For {$k = 1, \dots, K-1$}
		\State \textit{// 1. Subspace sketching based on the current subspace using Algorithm~\ref{alg:onelayer}}:
		\begin{align*}
		U_{[k]}^*, V_{[k]}^* 
		= \argmin\limits_{U_{[k]}, V_{[k]}} 
		\mathbb E_{(x, y)\sim \mathcal D} \left\{\ell(y, \relu \left(U_{[k]} V_{[k]} f_{[k-1]}(x)\right))\right\},   
		\end{align*}
		\State \textit{// 2. Expand the layer using the refined subspace as our new network}: $$f_{[k]} (x) = \relu \left(U_{[k]}^* V_{[k]}^* f_{[k-1]}(x)\right)$$
		\EndFor
		\State \Return $f= f_{[K]}$
	\end{algorithmic}
	\label{alg:subspace_expand}
\end{algorithm}

The single layer model in \eqref{eq:onelayer_model} has limited capability to 
capture the highly nonlinear regression relationships, as the parameters are 
linearly linked to the subspace except for a ReLU operation. However, the 
single-layer procedure in Algorithm \ref{alg:onelayer} has provided a building 
block, based on which we can develop an efficient algorithm to train a deep 
\textit{subspace network} (SN) in a greedy fashion. We thus propose a network 
expansion procedure to overcome such limitation.  

After we obtain the parameter subspace $U$ and sketch $V$ for the single-layer 
case \eqref{eq:onelayer_model}, we project the data points by $\bar x = 
\relu(UVx)$. A straightforward idea of the expansion is to use $(\bar x, y)$ as the new 
samples to train another layer. Let $f_{[k-1]}$ denote the network structure we 
obtained before the $k$-th expansion starts, $k = 1, 2, ..., K-1$, the expansion can 
recursively stack more ReLU layers:
\begin{equation} 
f_{[k]}(x) = \relu \left(U_{[k]} V_{[k]} f_{[k-1]}(x) + \epsilon \right),
\label{naivestack}
\end{equation}
However, we observe that simply stacking layers by repeating (\ref{naivestack}) 
many times can cause substantial information loss and degrade the generalization performance, 
especially since our training is layer-by-layer without ``looking back'' (i.e., top-down joint tuning). Inspired by deep residual 
networks \cite{he2016deep} that exploit ``skip connections'' to pass lower-level 
data and features to higher levels, we concatenate the original 
samples with the newly transformed, censored outputs after each time of expansion, i.e., reformulating 
$\bar x = [\relu(UVx); x]$ (similar manners could be found in \cite{zhou2017deep}). The new formulation after the expansion is given below:
\begin{align*}
f_{[k]}(x) = \relu \left(U_{[k]} V_{[k]} \left[f_{[k-1]}(x); x\right] + \epsilon \right).
\vspace{-1em}
\end{align*}


We summarize the network expansion process in 
Alg.~\ref{alg:subspace_expand}. The architecture of the resulting SN is illustrated in Fig.~\ref{fig:network_structure}. 
Compared to the single layer model \eqref{eq:onelayer_model}, SN gradually refines the parameter subspaces by multiple stacked nonlinear 
projections. It is expected to achieve superior performance due to the higher 
learning capacity, and the proposed SN can also be viewed as a gradient boosting method. Meanwhile, the layer-wise low-rank subspace structural prior would 
further improve generalization compared to naive multi-layer networks.

\section{Experiment}\label{sec:exp}
The subspace network code and scripts for generating 
the results in this section are available at 
\url{https://github.com/illidanlab/subspace-net}.

\subsection{Simulations on Synthetic Data}

\noindent \textbf{Subspace recovery in a single layer model.}
We first evaluate the 
subspace recovered by the proposed Algorithm~\ref{alg:onelayer} using synthetic data.
We generated $X\in \mathbb{R}^{N \times D}$, $U \in \mathbb{R}^{T \times R}$ and $V\in \mathbb{R}^{R \times D}$, all as i.i.d. random Gaussian matrices. The target matrix $Y\in \mathbb{R}^{N \times T}$ was then synthesized using (\ref{eq:onelayer_model}).
We set $N=5,000$, $D = 200$, $T=100$ $R=10$, and random noise as $\epsilon \sim \mathcal N(0,3^2)$. 

\begin{figure*}[t!]
	\centering
	\vspace{-2em}
	\subfloat[]{%
		{\includegraphics[scale=0.23]{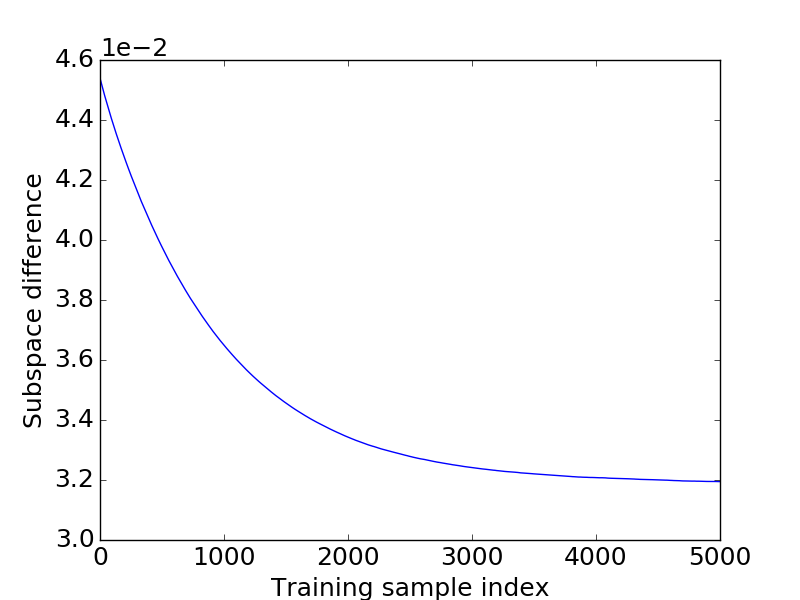}}%
		\label{U_diff}%
	}
	\hfill
	\subfloat[]{%
		{\includegraphics[scale=0.23]{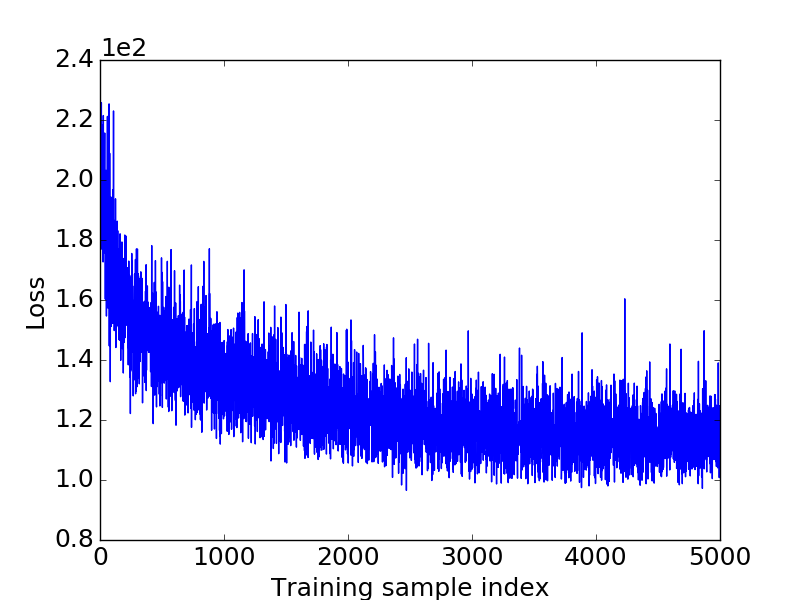}}%
		\label{online_conv}%
	}
	\hfill
	\subfloat[]{%
		{\includegraphics[scale=0.23]{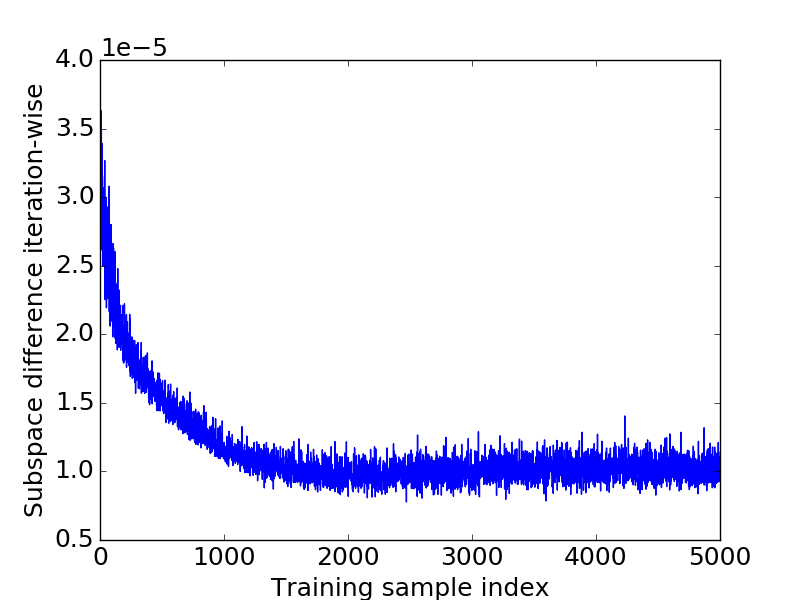}}%
		\label{U_iter}%
	}
	\vspace{-0.1in}
	\caption{Experimental results on subspace convergence. (a) Subspace differences, w.r.t. the index $i$; (b) Convergence of Algorithm~\ref{alg:onelayer}, w.r.t. the index $i$; (c) Iteration-wise subspace differences, w.r.t. the index $i$.}
	\label{fig:exp1_1}
	\vspace{-0.35in}
\end{figure*}

\begin{figure*}[t!]
	\centering
	\subfloat[]{%
		{\includegraphics[scale=0.23]{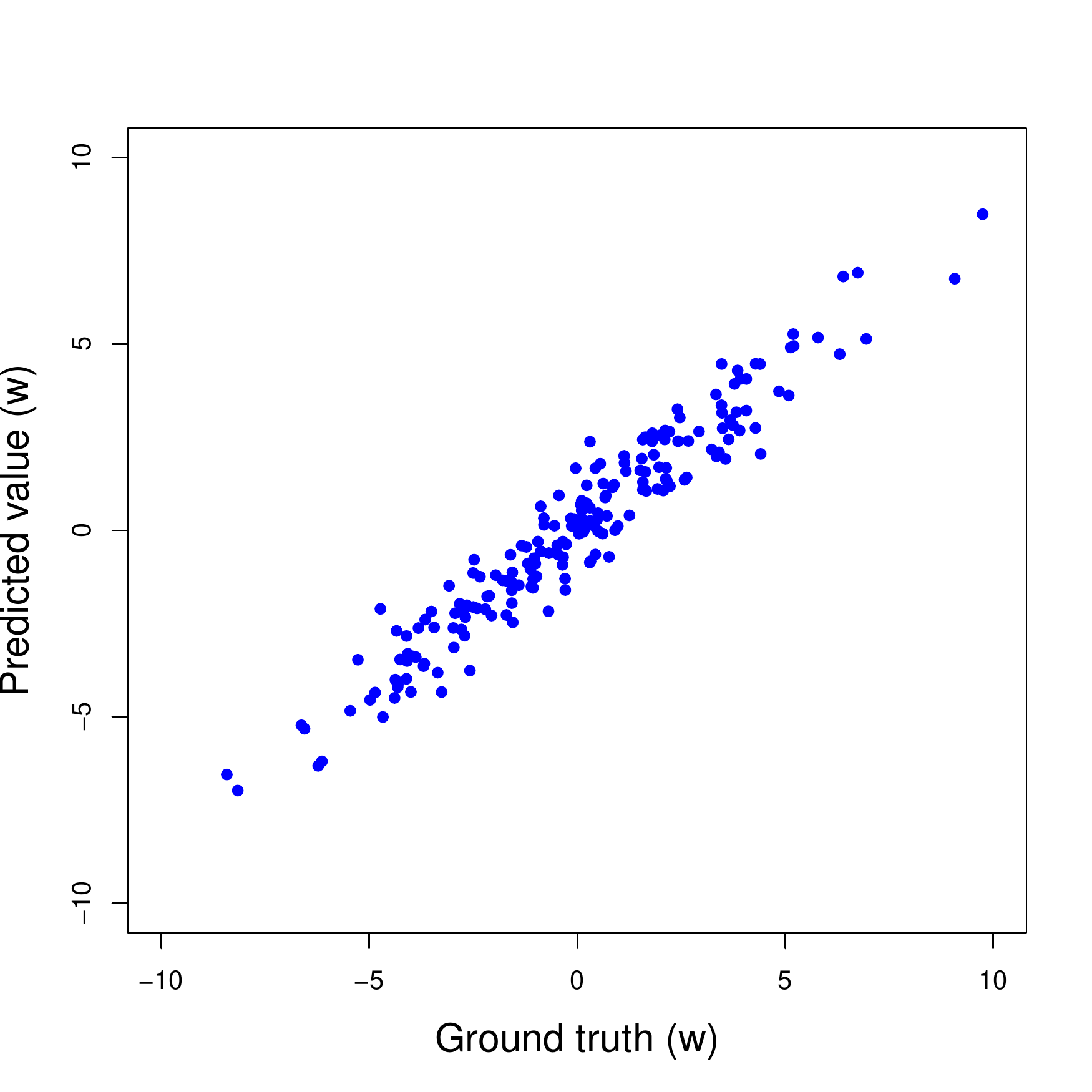}}%
		\label{task1}%
	}
	\hfill
	\subfloat[]{%
		{\includegraphics[scale=0.23]{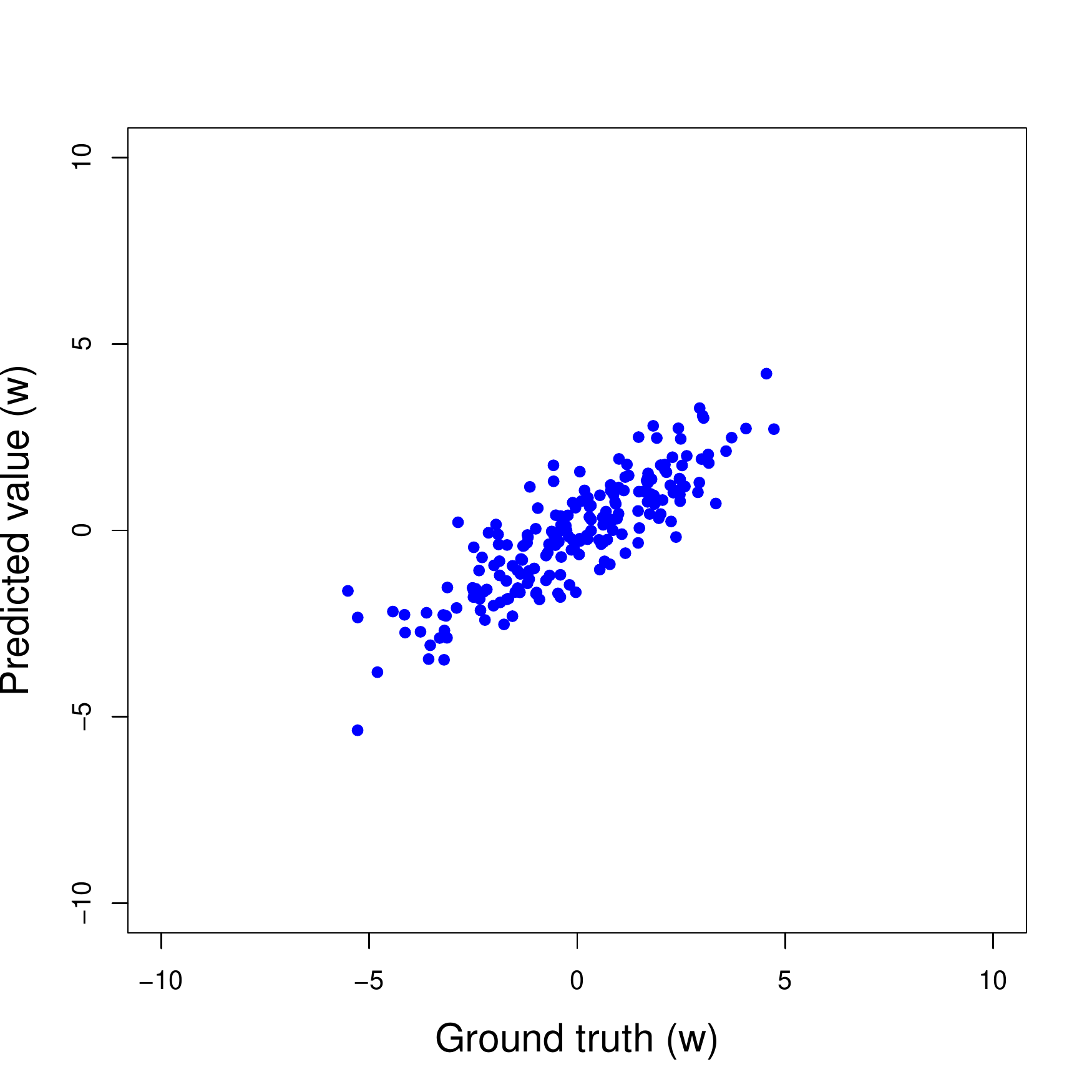}}%
		\label{task2}%
	}
	\hfill
	\subfloat[]{%
		{\includegraphics[scale=0.23]{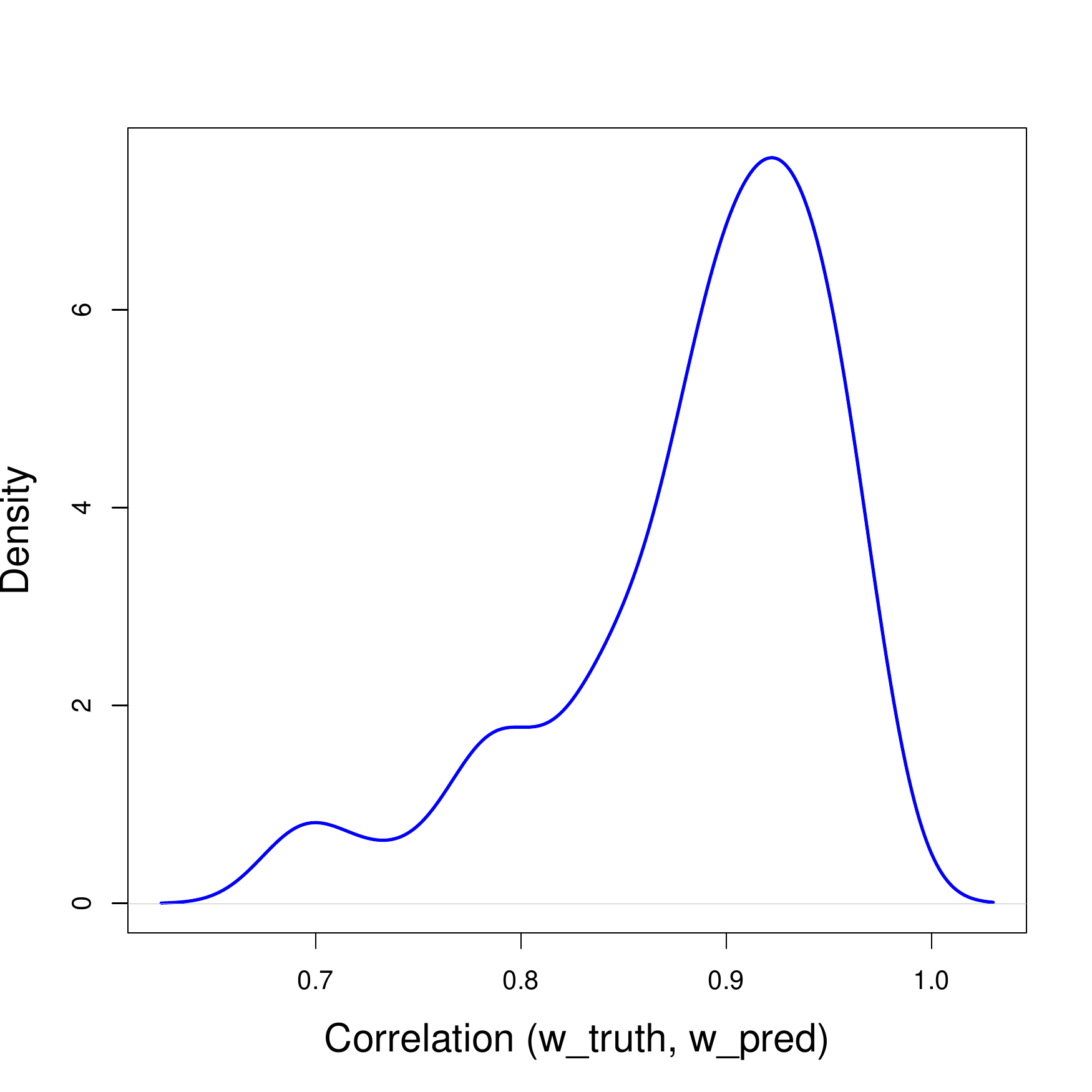}}%
		\label{density}%
	}
	\vspace{-0.1in}
	\caption{\small (a) Predicted weight vs true weight for task 1; (b) Predicted weight vs true weight for task 2; (c) Distribution of correlation between predicted weight and true weight for all tasks}
	\label{fig:single-layer}
	\vspace{-1em}
\end{figure*}

Figure~\ref{U_diff} shows the plot of \textit{subspace difference} between the ground-truth $U$ and the learned subspace $U_i$ throughout the iterations, i.e., $\| U-U_i\|_F/\|U\|_F$ w.r.t. $i$. This result verifies that Algorithm~\ref{alg:onelayer} is able to correctly find and smoothly converge to the underlying low-rank subspace of the synthetic data. The objective values throughout the online training process of Algorithm~\ref{alg:onelayer} are plotted in Figure~\ref{online_conv}. We further show the plot of \textit{iteration-wise subspace differences}, defined as $\|U_{i}-U_{i-1}\|_{F}/\|U\|_F$, in Figure~\ref{U_iter}, which complies with the $o(1/t)$ result in our non-asymptotic analysis. Moreover, the distribution of correlation between recovered weights and true weights for all tasks is given in Figure~\ref{fig:single-layer}, with most predicted weights having correlations with ground truth of above 0.9. 

\begin{table*}[t]
	\centering
	\caption{Comparison of subspace differences for each layer of SN, f-MLP, and rf-MLP.}
	\vspace{-1em}
	\label{tab:alg3_deep} 
		\begin{tabular}{|c||c|c|c||c|c|c||c|c|c|}
			\hline
			Metric  & \multicolumn{3}{c||}{Subspace Difference} & \multicolumn{3}{c||}{Maximum Mutual Coherence} & \multicolumn{3}{c|}{Mean Mutual Coherence} \\ \hline
			Method   & SN           & f-MLP       & rf-MLP      & SN         & f-MLP      & rf-MLP     & SN          & f-MLP      & rf-MLP     \\ \hline
			Layer 1 & \textbf{0.0313}       & 0.0315      & 0.0317      & 0.7608     & 0.7727     & \textbf{0.7895}     & \textbf{0.2900}      & 0.2725     & 0.2735     \\ \hline
			Layer 2 & 0.0321       & 0.0321      & 0.0321      & \textbf{0.8283}     & 0.7603     & 0.7654     & \textbf{0.2882}      & 0.2820     & 0.2829     \\ \hline
			Layer 3 & \textbf{0.0312}       & 0.0315      & 0.0313      & \textbf{0.8493}     & 0.7233     & 0.7890     & \textbf{0.2586}      & 0.2506     & 0.2485     \\ \hline
		\end{tabular}%
		\vspace{-0.15in}
\end{table*}

\vspace{+0.5em}
\noindent\textbf{Subspace recovery in a multi-layer subspace network.}
We re-generated synthetic data by repeatedly applying (\ref{eq:onelayer_model}) for three times, each time following the same setting as the single-layer model. A three-layer SN was then learned using Algorithm~\ref{alg:subspace_expand}. As one simple baseline, a \textit{multi-layer perceptron} (MLP) is trained, whose three hidden layers have the same dimensions as the three ReLU layers of the SN. Inspired by \cite{xue2013restructuring, sainath2013low, wang2015deepfont}, we then applied low-rank matrix factorization to each layer of MLP, with the same desired rank $R$, creating the \textit{factorized MLP} (f-MLP) baseline that has the identical architecture (including both ReLU hidden layers and linear bottleneck layers) to SN. We further re-trained the f-MLP on the same data from end to end, leading to another baseline, named \textit{retrained factorized MLP} (rf-MLP). 

Table~\ref{tab:alg3_deep} evaluates the subspace recovery fidelity in three layers, using three different metrics: (1) the maximum mutual coherence of all column pairs from two matrices, defined in \cite{candes2007sparsity} as a classical measurement on how correlated the two matrices' column subspaces are; (2) the mean mutual coherence of all column pairs from two matrices; (3) the subspace difference defined the same as in the single-layer case\footnote{The higher in terms of the two mutual coherence-based metrics, the better subspace recovery is achieved.That is different from the subspace difference case where the smaller the better, }. Note that the two mutual coherence-based metrics are immune to linear transformations of subspace coordinates, to which the $\ell_2$-based subspace difference might become fragile. SN achieves clear overall advantages under all three measurements, over f-MLP and rf-MLP. More notably, while the performance margin of SN in subspace difference seems to be small, the much sharper margins, in two (more robust) mutual coherence-based measurements, suggest that the recovered subspaces by SN are significantly better aligned with the groundtruth.

\begin{table*}[t!]
	\centering
	\caption{Average normalized mean square error under different approaches for synthetic data. 
	}
	\vspace{-0.1in}
	\label{sync-all} 
	\resizebox{0.8\textwidth}{!}{%
		\begin{tabular}{|c||c|c|c||c|c|}
			\hline
			\multirow{2}{*}{Percent} & \multicolumn{3}{c||}{Single Task (Shallow)} & \multicolumn{2}{c|}{Multi Task (Shallow)}        \\ \cline{2-6} 
			& Uncensored (LS + $\ell_1$)        & Censored (LS + $\ell_1$)          & Nonlinear Censored (Tobit)  & Uncensored (Multi Trace)          & Censored (Multi Trace)           \\ \hline
			40\%                     & 0.1412 (0.0007)  & \textbf{0.1127} (0.0010)  & \textbf{0.0428} (0.0003) & 0.1333 (0.0009)    & \textbf{0.1053} (0.0027)   \\ \hline
			50\%                     & 0.1384 (0.0005)  & \textbf{0.1102} (0.0010)  & \textbf{0.0408} (0.0004) & 0.1323 (0.0010)    & \textbf{0.1054} (0.0042)   \\ \hline
			60\%                     & 0.1365 (0.0005)  & \textbf{0.1088} (0.0009)  & \textbf{0.0395} (0.0003) & 0.1325 (0.0012)    & \textbf{0.1031} (0.0046)   \\ \hline
			70\%                     & 0.1349 (0.0005)  & \textbf{0.1078} (0.0010)  & \textbf{0.0388} (0.0004) & 0.1315 (0.0013)    & \textbf{0.1024} (0.0042)   \\ \hline
			80\%                     & 0.1343 (0.0011)  & \textbf{0.1070} (0.0012)  & \textbf{0.0383} (0.0006) & 0.1308 (0.0008)    & \textbf{0.1040} (0.0011)   \\ \hline \hline
			\multirow{2}{*}{Percent} & \multicolumn{3}{c||}{Deep Neural Network}              & \multicolumn{2}{c|}{Subspace Net (SN)} \\ \cline{2-6} 
			& DNN i (naive)        & DNN ii (censored)          & DNN iii (censored + low-rank)      & Layer 1            & Layer 3          \\ \hline
			40\%                     &  0.0623 (0.0041)   & 0.0489 (0.0035)  & \textbf{0.0431} (0.0041)  & 0.0390 (0.0004)    & \textbf{0.0369} (0.0002)   \\ \hline
			50\%                     &  0.0593 (0.0048)   & 0.0462 (0.0042) & \textbf{0.0400} (0.0039) & 0.0389 (0.0007)    & \textbf{0.0366} (0.0003)   \\ \hline
			60\%                     &  0.0587 (0.0053) & 0.0455 (0.0054) & \textbf{0.0395} (0.0050) & 0.0388 (0.0006)    & \textbf{0.0364} (0.0003)   \\ \hline
			70\%                     &  0.0590 (0.0071)  & 0.0447 (0.0043) & \textbf{0.0386} (0.0058) & 0.0388 (0.0006)    & \textbf{0.0363} (0.0003)   \\ \hline
			80\%                     &  0.0555 (0.0057)  & 0.0431 (0.0053) & \textbf{0.0380} (0.0057) & 0.0390 (0.0008)    &  \textbf{0.0364} (0.0005)   \\ \hline
		\end{tabular}%
	}
\end{table*}

\begin{table}[t!]
	\vspace{-1em}
	\centering
	\caption{Average normalized mean square error at each layer for subspace network ($R=10$) for synthetic data. 
	}
	\vspace{-0.5em}
	\Small
	\label{sync-subnet} 
		\begin{tabular}{|c|c|c|c|c|c|}
			\hline
			Perc. & Layer 1         & Layer 2         & Layer 3           & Layer 10        & Layer 20        \\ \hline\hline
			40\%    & 0.0390 (0.0004) & 0.0381 (0.0005) & 0.0369 (0.0002) & 0.0368 (0.0002) & \textbf{0.0368} (0.0002) \\ \hline
			50\%    & 0.0389 (0.0007) & 0.0379 (0.0005) & 0.0366 (0.0003)  & 0.0366 (0.0003) & \textbf{0.0365} (0.0003) \\ \hline
			60\%    & 0.0388 (0.0006) & 0.0378 (0.0004) & 0.0364 (0.0003)  & 0.0364 (0.0003) & \textbf{0.0363} (0.0003) \\ \hline
			70\%    & 0.0388 (0.0006) & 0.0378 (0.0005) & 0.0363 (0.0003) & 0.0363 (0.0003) & \textbf{0.0362} (0.0003) \\ \hline
			80\%    & 0.0390 (0.0008) & 0.0378 (0.0006) & 0.0364 (0.0005) & 0.0363 (0.0005) & \textbf{0.0363} (0.0005) \\ \hline
		\end{tabular}%
	\vspace{-2em}
\end{table}

\vspace{+0.5em}
\noindent \textbf{Benefits of Going Deep.} We re-generate synthetic data again in the same way as the \ul{first single-layer experiment}; yet differently, we now aim to show that a deep SN will boost performance over single-layer subspace recovery, even the data generation does not follow a known multi-layer model. We compare SN (both 1-layer and 3-layer) with two carefully chosen sets of state-of-art approaches: (1) single and multi-task ``shallow'' models; (2) deep models. 
For the first set, the least squares (LS) is treated as a naive baseline, while ridge (LS + $\ell_2$) and lasso (LS + $\ell_1$) regressions are considered for shrinkage or variables selection purpose; Censor regression, also known as the Tobit model, is a \textbf{non-linear} method to predict bounded targets , e.g., \cite{berberidis2016online}. Multi-task models with regularizations on trace norm (Multi Trace) and $\ell_{2,1}$ norm (Multi $\ell_{2,1}$) have been demonstrated to be successful on simultaneous structured/sparse learning, e.g., \cite{yang2010online, zhang2013robust}.\footnote{Least squares, ridge, lasso, and censor regression are implemented by \textsc{Matlab} optimization toolbox. MTLs are implemented through \textsc{MALSAR}~\cite{zhou2011malsar} with parameters carefully tuned.} We also verify \ul{the benefits of accounting for boundedness of targets} (Uncensored vs. Censored) in both single-task and multi-task settings, with \textbf{best} performance reported for each scenario (LS + $\ell_1$ for single-task and Multi Trace for multi-task).
For the set of deep model baselines, we construct three DNNs for fair comparison: i) A 3-layer fully connected DNN with the same architecture as SN, with a plain MSE loss; ii) A 3-layer fully connected DNN as i) with ReLU added for output layer before feeding into the MSE loss, which naturally implements non-negativity \textit{censored} training and evaluation; iii) A factorized and re-trained DNN from ii), following the same procedure of rf-MLP in the multi-layer synthetic experiment. Apparently, ii) and iii) are constructed to verify if DNN also benefits from the censored target and the low-rank assumption, respectively. 

We performed 10-fold random-sampling validation on the same dataset, i.e., randomly splitting into training and validation data 10 times. For each split, we fitted model on training data and evaluated the performance on validation data. Average normalized mean square error (ANMSE) across all tasks was obtained as the overall performance for each split. For methods without hyper parameters (least square and censor regression), an average of ANMSE for 10 splits was regarded as the final performance; for methods with tunable parameters, e.g., $\lambda$ in lasso, we performed a grid search on $\lambda$ values and chose the optimal ANMSE result. We considered different splitting sizes with training samples containing [40\%, 50\%, 60\%, 70\%, 80\%] of all the samples.  

Table \ref{sync-all} further compares the performance of all approaches. Standard deviation of 10 trials is given in parenthesis (same for all following tables). We can observe that: (1) all \textbf{censored} models significantly outperform their uncensored counterparts, verifying the necessity of adding censoring targets for regression. Therefore, we will use censored baselines hereinafter, unless otherwise specified; (2) the more structured \textbf{MTL} models tend to outperform single task models by capturing task relatedness. That is also evidenced by the performance margin of DNN iii over DNN i; (3) the \textbf{nonlinear} models are undoubtedly more favorable: we even see the single-task Tobit model to outperform MTL models; (4) As a \textit{nonlinear, censored MTL model}, SN combines the best of them all, accounting for its superior performance over all competitors. In particular, even a 1-layer SN already produces comparable performance to the 3-layer DNN iii (which also a nonlinear, censored MTL model trained with back-propagation, with three times the parameter amount of SN), thanks to SN's theoretically solid online algorithm in sketching subspaces. 

Furthermore, increasing the number of layers in SN from 2 to 20 demonstrated that SN can also benefit from growing depth without an end to end scheme. As Table~\ref{sync-subnet} reveals, SN steadily improves with more layers, until reaching a plateau at $\sim 5$ layers (as the underlying data distribution is relatively simple here). The observation is consistent among all splits.

\noindent\textbf{Computation speed.}
All experiments run on the same machine (1 x Six-core Intel Xeon E5-1650 v3 [3.50GHz], 12 logic cores, 128 GB RAM). GPU accelerations are enabled for DNN baselines, while \textbf{SN has not exploited the same accelerations yet}. 
The running time for a single round training on synthetic data (N=5000, D=200, T=100) is given in Table~\ref{run-time}. Training each layer of SN will cost 109 seconds on average. As we can see, SN improves generalization performance without a significant computation time burden. Furthermore, we can accelerate SN further, by reading data in batch mode and performing parallel updates.

\begin{table}[t!]
	\centering
	\vspace{-1em}
	\caption{
		Running time on synthetic data.}
	\vspace{-1em}
	\label{run-time} \small
	\begin{tabular}{|c|c|c|}
		\hline
		Method          & Time (s) & Platform \\ \hline
		\hline
		Least Square   & 0.02      & Matlab   \\ \hline
		LS+$\ell_2$     & 0.02      & Matlab   \\ \hline
		LS+$\ell_1$     & 18.4      & Matlab   \\ \hline
		Multi-trace     & 32.3       & Matlab   \\ \hline
		Multi-$\ell_{21}$ & 27.0       & Matlab   \\ \hline
		Censor          & 1680       & Matlab   \\ \hline
		SN (per layer)     & 109      & Python   \\ \hline
		DNN    & 659      & Tensorflow   \\ \hline
	\end{tabular}
	\vspace{-2em}
\end{table}

\begin{table*}[t!]
	\centering
	\caption{Average normalized mean square error at each layer for subspace network ($R=5$) for real data. 
	}
	\label{real-subnet} 
	\vspace{-0.18in}
	\small
	\begin{tabular}{|c|c|c|c|c|c|}
		\hline
		Percent & Layer 1         & Layer 2         & Layer 3         & Layer 5         & Layer 10        \\ \hline
		40\%    & 0.2016 (0.0057) & 0.2000 (0.0039) & 0.1981 (0.0031) & 0.1977 (0.0031) & \textbf{0.1977} (0.0031)  \\ \hline
		50\%    & 0.1992 (0.0040) & 0.1992 (0.0053) & 0.1971 (0.0038) & 0.1968 (0.0036) & \textbf{0.1967} (0.0035) \\ \hline
		60\%    & 0.1990 (0.0061) & 0.1990 (0.0047) & 0.1967 (0.0038) & 0.1964 (0.0039) & \textbf{0.1964} (0.0038) \\ \hline
		70\%    & 0.1981 (0.0046) & 0.1966 (0.0052) & 0.1953 (0.0039) & 0.1952 (0.0039) & \textbf{0.1951} (0.0038) \\ \hline
		80\%    & 0.1970 (0.0034) & 0.1967 (0.0044) & 0.1956 (0.0040) & 0.1955 (0.0039) & \textbf{0.1953} (0.0039) \\ \hline
	\end{tabular}%
\end{table*}

\begin{table*}[tbh!]
	\vspace{-1em}
	\centering
	\caption{Average normalized mean square error under different approaches for real data. 
	}
	\label{real-all} \scriptsize
	\vspace{-1em}
	\resizebox{0.8\textwidth}{!}{%
	\begin{tabular}{|c||c|c|c||c|c|}
		\hline
		\multirow{2}{*}{Percent} & \multicolumn{3}{c||}{Single Task (Censored)} & \multicolumn{2}{c|}{Multi Task (Censored)}   \\ \cline{2-6} 
		& Least Square        & LS + $\ell_1$       & Tobit (Nonlinear)    & Multi Trace     & Multi $\ell_{2,1}$       \\ \hline
		40\%                     & 0.3874 (0.0203)     & \textbf{0.2393} (0.0056)    & 0.3870 (0.0306)    & 0.2572 (0.0156) & \textbf{0.2006} (0.0099) \\ \hline
		50\%                     & 0.3119 (0.0124)     & \textbf{0.2202} (0.0049)    & 0.3072 (0.0144)    & 0.2406 (0.0175) & \textbf{0.2002} (0.0132) \\ \hline
		60\%                     & 0.2779 (0.0123)     & \textbf{0.2112} (0.0055)    & 0.2719 (0.0114)    & 0.2596 (0.0233) & \textbf{0.2072} (0.0204) \\ \hline
		70\%                     & 0.2563 (0.0108)     & \textbf{0.2037} (0.0042)    & 0.2516 (0.0108)    & 0.2368 (0.0362) & \textbf{0.2017} (0.0116) \\ \hline
		80\%                     & 0.2422 (0.0112)     & \textbf{0.2005} (0.0054)    & 0.2384 (0.0099)    & 0.2176 (0.0171) & \textbf{0.2009} (0.0050) \\ \hline \hline
		\multirow{2}{*}{Percent} & \multicolumn{3}{c||}{Deep Neural Network}                      & \multicolumn{2}{c|}{Subspace Net (SN)}           \\ \cline{2-6} 
		& DNN i (naive)           & DNN ii (censored)          & DNN iii (censored + low-rank)         & Layer 1         & Layer 3        \\ \hline 
		40\%                     & 0.2549 (0.0442)     & 0.2388 (0.0121)    & \textbf{0.2113} (0.0063)    & 0.2016 (0.0057) & \textbf{0.1981} (0.0031)  \\ \hline
		50\%                     & 0.2236 (0.0066)     & 0.2208 (0.0062)    & \textbf{0.2127} (0.0118)    & 0.1992 (0.0040) & \textbf{0.1971} (0.0038) \\ \hline
		60\%                     & 0.2215 (0.0076)     & 0.2200 (0.0076)    & \textbf{0.2087} (0.0102)    & 0.1990 (0.0061) & \textbf{0.1967} (0.0038) \\ \hline
		70\%                     & 0.2149 (0.0077)     & 0.2141 (0.0079)    & \textbf{0.2093} (0.0137)    & 0.1981 (0.0046) & \textbf{0.1953} (0.0039) \\ \hline
		80\%                     & 0.2132 (0.0138)     & 0.2090 (0.0079)    & \textbf{0.2069} (0.0135)    & 0.1970 (0.0034) & \textbf{0.1956} (0.0040) \\ \hline
	\end{tabular}%
}
	\vspace{-1em}
\end{table*}

\begin{table*}[t!]
	\centering
	\caption{Average normalized mean square error under different rank assumptions for real data. 
	}
	\label{real-rank}
	\vspace{-0.1in}
		 \small
		\begin{tabular}{|c|c|c|c|c|c|}
			\hline
			Method                                                                                   & Percent - Rank & $R = 1$              & $R = 3$             & $R = 5$             & $R = 10$            \\ \hline
			\multirow{5}{*}{SN}                                                                      & 40\%           & 0.2052 (0.0030)  & 0.1993 (0.0036) & \textbf{0.1981} (0.0031) & 0.2010 (0.0044) \\ \cline{2-6} 
			& 50\%           & 0.2047 (0.0029)  & 0.1983 (0.0034) & \textbf{0.1971} (0.0038) & 0.2001 (0.0046) \\ \cline{2-6} 
			& 60\%           & 0.2052 (0.0033)  & 0.1988 (0.0047) & \textbf{0.1967} (0.0038) & 0.1996 (0.0052) \\ \cline{2-6} 
			& 70\%           & 0.2043 (0.0044)  & 0.1975 (0.0042) & \textbf{0.1953} (0.0039) & 0.1990 (0.0057) \\ \cline{2-6} 
			& 80\%           & 0.2058 (0.0051)  & 0.1977 (0.0042) & \textbf{0.1956} (0.0040) & 0.1990 (0.0058) \\ \hline
			\multirow{5}{*}{\begin{tabular}[c]{@{}c@{}}DNN iii\\ (censored + low-rank)\end{tabular}} & 40\%           & 0.2322 (0.0146)  & 0.2360 (0.0060) & \textbf{0.2113} (0.0063) & 0.2196 (0.0124) \\ \cline{2-6} 
			& 50\%           & 0.2298 (0.0093)  & 0.2256 (0.0127) & \textbf{0.2127} (0.0118) & 0.2235 (0.0142) \\ \cline{2-6} 
			& 60\%           & 0.2244 (0.0132)  & 0.2277 (0.0099) & \textbf{0.2087} (0.0102) & 0.2145 (0.0208) \\ \cline{2-6} 
			& 70\%           & 0.2178 (0.0129)  & 0.2177 (0.0115) & \textbf{0.2093} (0.0137) & 0.2083 (0.0127) \\ \cline{2-6} 
			& 80\%           & 0.2256 (0.0117) & 0.2250 (0.0079) & \textbf{0.2069} (0.0135) & 0.2158 (0.0183) \\ \hline
		\end{tabular}%
		\vspace{-0.1in}
\end{table*}

\begin{table}[t!]
	\centering
	\vspace{-0.5em}
	\caption{\small 
	Average normalized mean square error for non-calibrated vs. calibrated SN for real data (6 layers). 
	}
	\vspace{-0.5em}
	\label{cal-noncal}
	\small
	\begin{tabular}{|c|c|c|c|c|}
		\hline
		Percent & Non-calibrate          & Calibrate         \\ \hline
		40\%         & 0.1993 (0.0034) & \textbf{0.1977} (0.0031) \\ \hline
		50\%         & 0.1987 (0.0043) & \textbf{0.1967} (0.0036)  \\ \hline
		60\%         & 0.1991 (0.0044) & \textbf{0.1964} (0.0039)  \\ \hline
		70\%         & 0.1982 (0.0042) & \textbf{0.1951} (0.0038)  \\ \hline
		80\%         & 0.1984 (0.0041) & \textbf{0.1954} (0.0039)  \\ \hline
	\end{tabular}
	\vspace{-1.5em}
\end{table}

\vspace{-1em}
\subsection{Experiments on Real data}

We evaluated SN in a real clinical setting to build
models for the prediction of important clinical scores representing a subject's
cognitive status and signaling the progression of Alzheimer's disease (AD), from structural Magnetic Resonance
Imaging (sMRI) data. AD is one major neurodegenerative disease that accounts for 60 to 80 percent of dementia. The National Institutes of Health has thus focused on studies investigating brain and fluid biomarkes of the disease, and supported the long running project Alzheimer's Disease Neuroimaging Initiative (ADNI) from 2003. We used the ADNI-1 cohort (\url{http://adni.loni.usc.edu/}). In the experiments, we used the 1.5 Tesla structural MRI collected at the baseline, and performed
cortical reconstruction and volumetric segmentations with the FreeSurfer
following the procotol in~\cite{jack2008alzheimer}. For each MRI image, we
extracted 138 features representing the cortical thickness and surface areas
of region-of-interests (ROIs) using the Desikan-Killiany cortical
atlas~\cite{desikan2006automated}. After preprocessing, we obtained a dataset
containing 670 samples and 138 features. These imaging features were used to
predict a set of 30 clinical scores including ADAS scores~\cite{rosen1984new}
at baseline and future (6 months from baseline), baseline Logical Memory from
Wechsler Memory Scale IV~\cite{scale2009edition}, Neurobattery scores (i.e.
immediate recall total score and Rey Auditory Verbal Learning Test scores),
and the Neuropsychiatric Inventory~\cite{cummings1997neuropsychiatric} at
baseline and future.

\noindent\textbf{Calibration.} 
In MTL formulations we typically assume that noise variance $\sigma^2$ is the same across all tasks, which may not be true in many cases. To deal with heterogeneous $\sigma^2$ among tasks, we design a \textit{calibration} step in our optimization process, where we estimate task-specific $\hat{\sigma}_t^2$ using $\|y-\hat{y}\|_2^2/N$ before ReLU, as the input for next layer and repeat on layer-wise. We compare performance of both non-calibrated and calibrated methods.

\noindent\textbf{Performance.} 
We adopted the two sets of baselines used in the last synthetic experiment for the real world data. Different from synthetic data where the low-rank structure was predefined, for real data, there is no groundtruth rank available and we have to try different rank assumptions. Table~\ref{cal-noncal} compares the performances between $\sigma^2$ non-calibrated versus calibrated models. We observe a clear improvement by assuming different $\sigma^2$ across tasks. Table~\ref{real-all} shows the results for all comparison methods, with SN outperforming all else. Table~\ref{real-subnet} shows the SN performance growth with increasing the number of layers. Table~\ref{real-rank} further reveals the performance of DNNs and SN using varying rank estimations in real data. As expected, the U-shape curve suggests that an overly low rank may not be informative enough to recover the original weight space, while a high-rank structure cannot enforce as strong a structural prior. However, the overall robustness of SN to rank assumptions is fairly remarkable: its performance under all ranks is competitive, consistently outperforming DNNs under the same rank assumptions and other baselines.

\begin{figure}[t!]
\centering 
\includegraphics[width=0.45\textwidth]{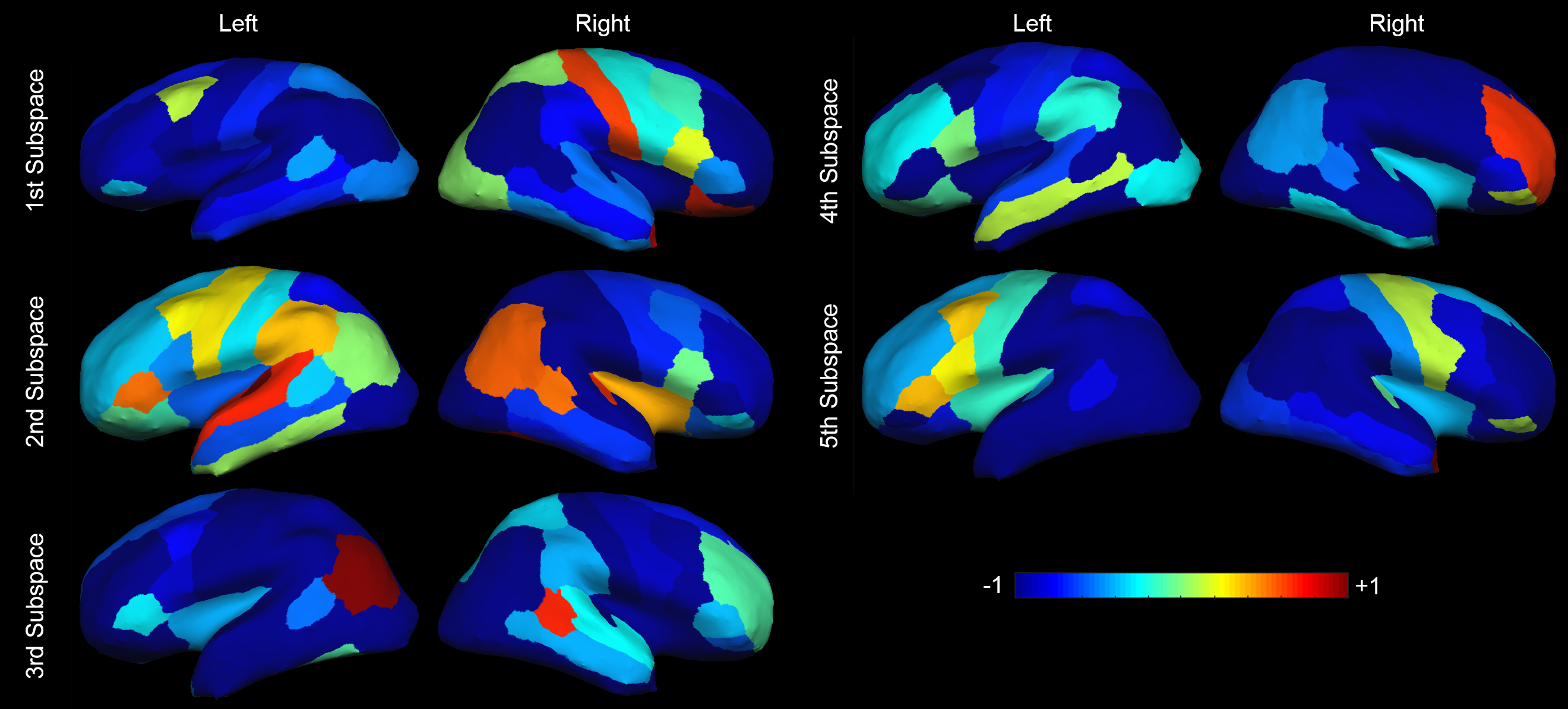}
\vspace{-0.1in}
\caption{Brain mapping of 5 lowest-level subspaces identified in the proposed Subspace Network.}
\label{fig:brain_map_v1}
\vspace{-0.2in}
\end{figure}

\noindent\textbf{Qualitative Assessment.}
From the multi-task learning perspective, the subspaces serve as the shared
component for transferring predictive knowledge among the censored learning
tasks. The subspaces thus capture important predictive information in
predicting cognitive changes. We normalized the magnitude of the subspace into
the range of $[-1, 1]$ and visualized the subspace in brain mappings. The the 5
lowest level subspaces in $V_1$ are the most important five subspaces, and is
illustrated in Figure~\ref{fig:brain_map_v1}. 

We find that each subspace captures very different information. 
In the first subspace, the volumes of right \emph{banks of the superior temporal sulcus},
which is found to involve in prodromal AD~\cite{killiany2000use}, 
\emph{rostral middle frontal gyrus}, 
with highest A$\beta$ loads in AD pathology~\cite{nicoll2003neuropathology},
and the volume of \emph{inferior parietal lobule},
which was found to have an increased 
\emph{S-glutathionylated} proteins in a proteomics study~\cite{newman2007increase}, 
have significant magnitude. We also find evidence of strong 
association between AD pathology and brain regions of large magnitude in 
other subspaces. The subspaces in remaining
levels and detailed clinical analysis will be available in a journal 
extension of this paper.



\section{Conclusions and future work}
In this paper, we proposed a {\it Subspace Network} (SN), an efficient
deep modeling approach for non-linear multi-task censored regression, where each
layer of the subspace network performs a multi-task censored regression to
improve upon the predictions from the last layer via sketching a low-dimensional subspace to perform knowledge transfer among learning tasks. We show that under mild assumptions, for each layer we can recover the parametric
subspace using only one pass of training data. We demonstrate empirically that the subspace network can quickly capture correct parameter subspaces, and outperforms state-of-the-arts in predicting neurodegenerative clinical scores from brain imaging. Based on similar formulations, the proposed method can be easily extended to cases where the targets have nonzero bounds, or both lower and upper bounds.

\section*{Appendix} 

We hereby give more details for the proofs of both asymptotic and non-asymptotic convergence properties for 
Algorithm \ref{alg:onelayer} to recover the latent subspace $U$. The proofs heavily rely on a series of previous results in~\cite{mairal2010online, kasiviswanathan2012online, 
	shalev2012online, mardani2013dynamic, mardani2015subspace, shen2016online}, and many key results are directly referred to hereinafter for conciseness. We include the proofs for the manuscript to be self-contained. 

At iteration $i = 1, 2,..., N$,  we sample $(x_i, y_i)$, and let $U^i, V^i$ denote the intermediate $U$ and $V$, to be differentiated from $U_t, V_t$ which are the $t$-th columns of $U, V$. For the proof feasibility, we assume that $\{(x_i, y_i)\}_{i=1}^N$ are sampled i.i.d., and the subspace sequence $\{U^i\}_{i=1}^N$ lies in a compact set. 

\subsection*{Proof of Asymptotic Properties}

For {\it infinite} data streams with $N \rightarrow \infty$, we recall the instantaneous cost of the $i$-th datum: 
\begin{align*}
g_i(x_i, y_i, U, V)  =  - \log\Pr(x_i, y_i | U, V) +  \frac{\lambda}{2}\|U\|_F^2 + 
\frac{\lambda}{2}\|V\|_F^2,
\end{align*} 
and the online optimization form recasted as an empirical cost  minimization:
\begin{align*}
\min\nolimits_{U} \frac{1}{N}\sum\nolimits_{i=1}^N g_i(x_i, y_i, U, V).
\end{align*}
The Stochastic Gradient Descent (SGD) iterations can be seen as  minimizing the approximate cost:
\begin{align*}
\min\nolimits_{U} \frac{1}{N}\sum\nolimits_{i=1}^N g'_i(x_i, y_i, U, V).
\end{align*}
where $g'_N$ is a tight quadratic surrogate for $g_N$ based on the second-order Taylor approximation around $U^{N-1}$:
\begin{align*}
g'_N(x_N, y_N, U, V) &= g_N(x_N, y_N, U^{N-1}, V) 
\\&+ \langle \nabla_{U} g_N(x_N, y_N, U^{N-1}, V), U-U^{N-1} \rangle 
\\&+ \frac{\alpha_N}{2} \|U-U^{N-1}\|_F^2,
\end{align*}
with $\alpha_N \geq \|\nabla^2_{U} g_N(x_N, y_N, U^{N-1},V)\|$. $g'_N$ is further recognized as a locally tight upper-bound surrogate for $g_N$, with locally tight gradients. Following the Appendix 1 of~\cite{shen2016online}, we can show that $g_N$ is smooth, with its first-order and second-order gradients bounded w.r.t. each $U_N$. 

With the above results, the convergence of subspace iterates can be proven in the same regime developed in \cite{mardani2015subspace}, whose main inspirations came from  \cite{mairal2010online} that established convergence of an online dictionary learning algorithm using the martingale sequence theory. In a nutshell, the proof procedure proceeds by first showing that $\sum_{i=1}^N g_i'(x_i, y_i, U^i, V^i)$ converges to $\sum_{i=1}^N g_i(x_i, y_i, U^i, V^i)$ asymptotically, according to the 
quasi-martingale property in the almost sure sense, owing to the tightness of $g'$. It then implies convergence of the associated gradient 
sequence, due to the regularity of $g$. 

Meanwhile, we notice that $g_i(x_i, y_i, U, V)$ is bi-convex for the block variables $U_t$ and $V$ (see Lemma 2 of ~\cite{shen2016online}). Therefore due to the convexity of $g_N$ w.r.t. $V$ when $U = U^{N-1}$ is fixed, the parameter sketches $V$ can also be updated exactly per iteration. 

All above combined, we can claim the asymptotic convergence  for the iterations of Algorithm \ref{alg:onelayer}: as $N \rightarrow \infty$, \textit{the subspace  sequence $\{U^i\}_{i=1}^N$ asymptotically converges to a stationary-point of the 
	batch estimator}, under a few mild conditions.

\subsection*{Proof of Non-Asymptotic Properties}

For {\it finite} data streams, we rely on the unsupervised formulation of regret analysis ~\cite{kasiviswanathan2012online, 
	shalev2012online} to assess the performance of online iterates. Specifically, at iteration $t$ ($t \le N$), we use the previous
$U^{t-1}$ to span the partial data at $i = 1, 2, ..., t$. Prompted by the alternating nature of iterations, we adopt a variant of the unsupervised regret to assess the goodness of online subspace estimates in representing the partially available data. With $g_t(x_t, y_t, U^{t-1}, V)$ being the loss incurred by the estimate $U^{t-1}$ for predicting the $t$-th datum, the cumulative online loss for a stream of size $T$ is given by:
\begin{align}
\bar{C}_T:= \frac{1}{T}\sum_{\tau=1}^T g_{\tau}(x_{\tau}, y_{\tau},U^{\tau-1}, V).\label{eq:c:bar}
\end{align}
Further, we will assess the cost of the last estimate $U^T$ using:
\begin{align}
\hat{C}_T= \frac{1}{T}\sum_{\tau=1}^T g_{\tau}(x_{\tau}, y_{\tau},U^T, V).\label{eq:avg_online_loss}
\end{align}
We define $C_T$ as the batch estimator cost. For the sequence $\{U^t\}_{t=1}^T$, we define the online regret: 
\begin{align}
\mathcal{R}_T:= \hat{C}_T - \bar{C}_T.    \label{def:regret}
\end{align}
We investigate the convergence rate of the sequence $\{\mathcal{R}_T\}$ to zero as $T$ grows. Due to the nonconvexity of the online subspace iterates, it is challenging to directly analyze how fast the online cumulative loss $\bar{C}_t$ approaches the optimal batch cost $C_t$. As \cite{shen2016online} advocates, we instead investigate whether $\hat{C}_t$ converges to $\bar{C}_t$. That is established by first referring to the Lemma 2 of~\cite{mardani2013dynamic}: \textit{the distance  between successive subspace estimates will vanish as fast as $O(1/t)$: $\|U^t -  U^{t-1}\|_F \le \frac{B}{t}$, for some constant $B$ that is independent of $t$  and $N$}. 
Following the proof of Proposition 2 in~\cite{shen2016online}, we can similarly show that: if $\{U^t\}_{t=1}^T$ and $\{V_tx_t\}_{t=1}^T$ are uniformly bounded, i.e.,~ $\|U^t\|_F\leq B_1$, and $\|V_tx_t\|_2\leq B_2$, $\forall t \le T$, then with constants $B_{1}, B_{2} >0$ and by choosing a constant step size $\mu_t=\mu$, we have a bounded regret as:
\begin{align}
\mathcal{R}_T\leq&  \frac{B^2(\ln(T)+1)^2}{2\mu T} +\frac{5 B^2}{6\mu T}.\nonumber
\end{align}


This thus concluded the proof.


\begin{acks}
This research is supported in part by National Science Foundation under Grant IIS-1565596, IIS-1615597, the Office of Naval Research under grant number N00014-17-1-2265, N00014-14-1-0631. 
\end{acks}

\bibliographystyle{ACM-Reference-Format}
\bibliography{reference}

\end{document}